\documentclass[letterpaper, 10 pt, journal, twoside]{IEEEtran}  % Comment this line out if you need a4paper

\IEEEoverridecommandlockouts                              % This command is only needed if 
\usepackage[margin=1in]{geometry}
\usepackage{amsmath} % assumes amsmath package installed
\usepackage{amssymb}  % assumes amsmath package installed
\usepackage{verbatim}
\usepackage[pdf]{pstricks}
\usepackage{tikz}
\usepackage{breqn}
\usepackage{booktabs} %for nice table layout
\usepackage[utf8]{inputenc}
\usepackage{algpseudocode}
\usepackage{algorithm}
\usepackage{url}
\usepackage{hyperref}
\usepackage{xcolor}
\usepackage{pgfplots}
\usepackage{graphicx}
\usepackage{caption}
\usepackage{subcaption}
\usepgfplotslibrary{groupplots}
\usetikzlibrary{fit,positioning}
\usetikzlibrary{patterns}
\usetikzlibrary{bayesnet}

\title{\LARGE \bf
Real Time Trajectory Prediction Using Deep Conditional Generative Models
}

\author{Sebastian Gomez-Gonzalez$^{1,2}$, Sergey Prokudin$^{1}$, Bernhard Sch\"olkopf$^{1}$ and Jan Peters$^{2}$% <-this % stops a space
\thanks{$^{1}$Max Planck for Intelligent Systems,
        Max Planck Ring 4, 72072 Tubingen, Germany
        {\tt\small sebastian@robot-learning.de}}%
\thanks{$^{2}$Technische Universitaet Darmstadt,
        Hochschulstrasse 27, 64289 Darmstadt, Germany
        {\tt\small mail@jan-peters.net}}%
\thanks{Digital Object Identifier (DOI): see top of this page.}
}

\newcommand{\vect}[1]{\boldsymbol{#1}}
\newcommand{\matr}[1]{\boldsymbol{#1}}
\newcommand{\trans}[1]{{#1}^{\top}}
\newcommand{\given}[0]{\, | \,}
\newcommand{\normal}[3]{\mathcal{N}\left({#1} \, \left| \, {#2},{#3}\right.\right)}

\DeclareMathOperator{\KL}{KL}
\DeclareMathOperator{\E}{\mathbb{E}}

\newcommand{\obs}[1]{\vect{y}_{#1}}
\newcommand{\yv}[1]{\vect{y}_{#1}} % y random variable
\newcommand{\yest}[0]{\hat{\vect{y}}}
\newcommand{\invar}[0]{\vect{x}}
\newcommand{\inmask}[0]{\hat{\vect{x}}} % unscripted mask variable
\newcommand{\zv}[0]{\vect{z}}
\newcommand{\ff}[0]{\vect{f}_\theta}
\newcommand{\gf}[0]{\vect{g}_\phi}

\newlength\figureheight
\newlength\figurewidth
\newlength\axislabelsep

\newcommand{\rev}[1]{#1}

\begin{document}

\maketitle
\thispagestyle{empty}
\pagestyle{empty}

%%%%%%%%%%%%%%%%%%%%%%%%%%%%%%%%%%%%%%%%%%%%%%%%%%%%%%%%%%%%%%%%%%%%%%%%%%%%%%%%
\begin{abstract}
  Data driven methods for time series forecasting that quantify uncertainty open 
  new important possibilities for robot tasks with hard real time constraints,
  allowing the robot system to make decisions that trade off between reaction time 
  and accuracy in the predictions. Despite the recent advances in deep learning,
  it is still challenging to make long term accurate predictions with the low
  latency required by real time robotic systems.
  In this paper, we propose a deep conditional generative model for trajectory prediction 
  that is learned from a data set of collected trajectories. Our method uses encoder and
  decoder deep networks that map complete or partial trajectories to a Gaussian distributed
  latent space and back, allowing for fast inference of the future values of a trajectory
  given previous observations. The encoder and decoder networks are trained using stochastic
  gradient variational Bayes.
  In the experiments, we show that our model provides more accurate
  long term predictions with a lower latency than popular models for trajectory forecasting
  like recurrent neural networks or physical models based on differential equations. Finally,
  we test our proposed approach in a robot table tennis scenario to evaluate the performance
  of the proposed method in a robotic task with hard real time constraints.
\end{abstract}


\section{INTRODUCTION}

\IEEEPARstart{D}{ynamic} high speed robotics tasks often require accurate methods to
forecast the future value of a physical quantity based on previous measurements
while respecting the real time constraints of the particular application. 
For example, to hit or catch a flying ball with a robotic system we need to predict 
accurately and fast the trajectory of the ball based on previous observations that 
are often noisy and might include outliers or missing observations. Note that the
time it takes to compute the predictions, called latency, is as important for
the application as the accuracy in the prediction. In our previous example, the
prediction of the future ball positions are only useful if the computation time
is significantly faster than the ball itself.

\begin{figure}[!b]
  \centering
  \includegraphics[width=7cm]{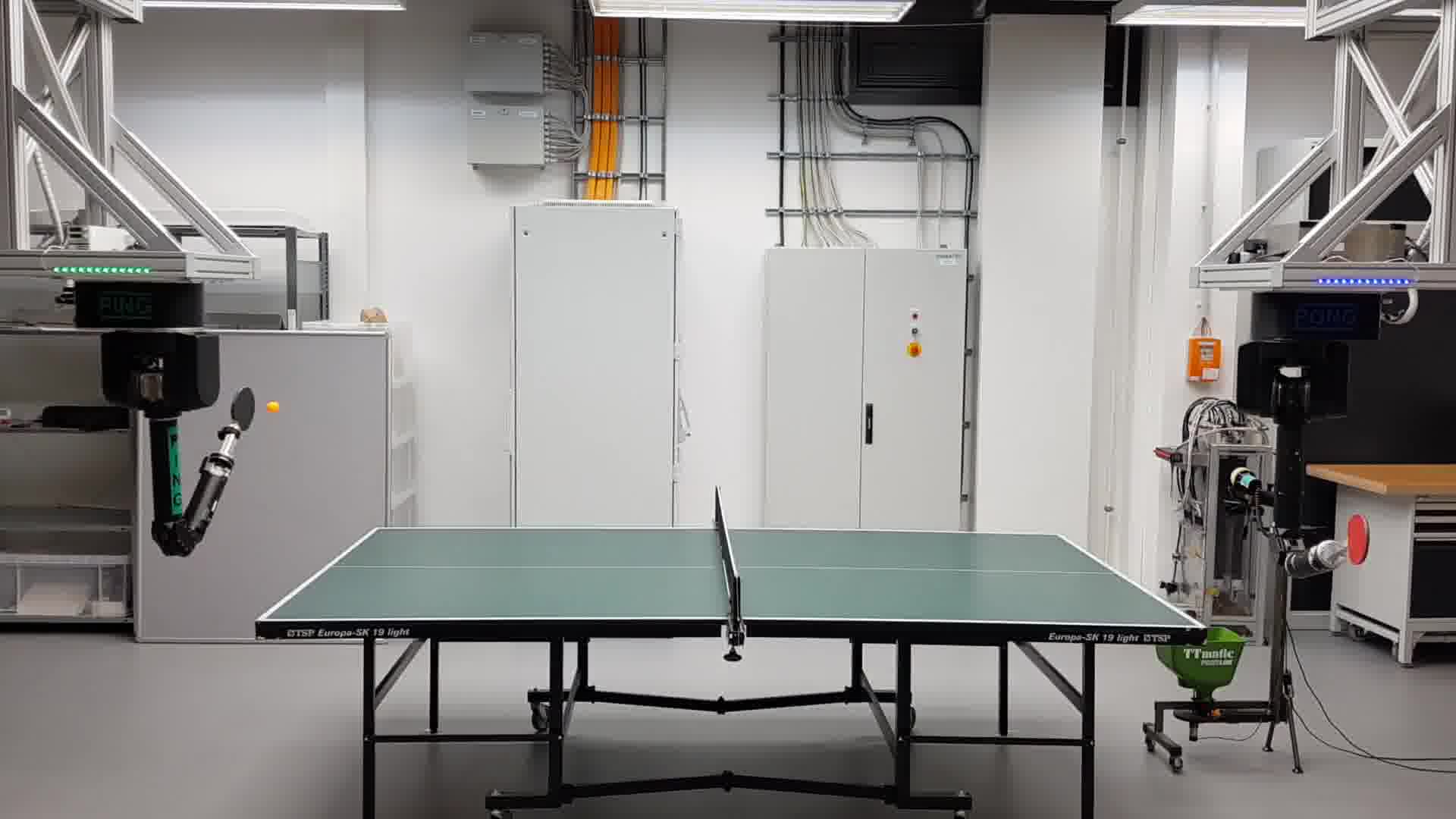}
  \caption{
    Robot table tennis setup used to evaluate the trajectory forecasting methods. 
    The ball is tracked using four VGA resolution cameras attached to the ceiling
    with a sampling frequency of 180 frames per second. The robot arms are
    Barrett WAM capable of high speed motion with seven degrees of freedom.
  }
  \label{fig:robot}
\end{figure}

Both physics-based~\cite{zhao2017model} and data-driven~\cite{cnnforecasting} 
models are used for trajectory forecasting. Physical models based on differential 
equations have been typically preferred to model and predict trajectories in 
high speed robotic systems~\cite{mulling2011biomimetic}, because they are 
relatively fast for predictions and are well studied models known to provide 
reasonably good predictions for many problems. However,
in some applications like pneumatic muscle robots~\cite{buchler2016lightweight}, 
the best known physic-based models are not accurate enough to be useful for 
control. Even in cases where the physics are relatively well known, estimating 
all the relevant variables to model the system can be difficult. In table tennis, 
for example, estimating the spin of the ball in real time from images is hard. 
In addition, small lens distortion on the vision system makes the position 
estimates not equally accurate in all the robot work space, rendering the 
estimation of the initial position and velocity less accurate.
A data-driven approach, on the other hand, may have the potential to estimate
the spin from its effect on the trajectory and ignore the lens distortion as
long as it is present both at training and test time. However, popular
data-driven methods for time series modeling like recurrent neural 
networks~\cite{pixelrnn} and auto-regressive models~\cite{cnnforecasting}
suffer from cumulative errors that render trajectory forecasting inaccurate
as we predict farther into the future.

\begin{figure*}
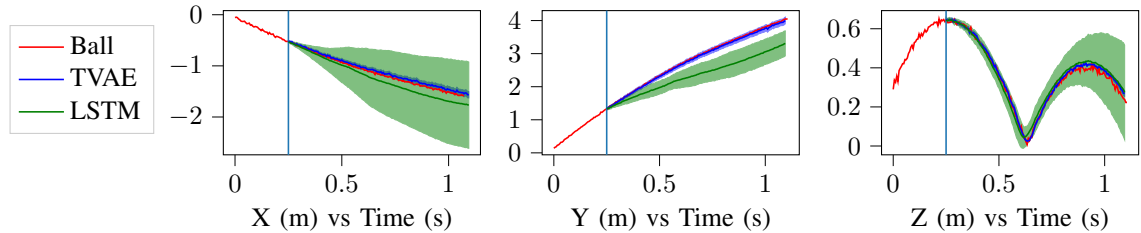

  \centering
  \setlength\figurewidth{5cm}
  \setlength\figureheight{3.5cm}
  \setlength\axislabelsep{0mm}
  \input{fig/d0_given45}
  \input{fig/d1_given45}
  \input{fig/d2_given45}
  \caption{
    Example of a ball trajectory in X, Y and Z (in meters) with respect to time 
    (in seconds) and the respective prediction using LSTMs and the proposed 
    method (TVAE).
    The observed ball trajectory is depicted in red, the prediction using a LSTM
    is depicted in green, and the prediction using the proposed model is depicted
    in blue. The shaded area corresponds to one standard deviation. 
    Note the cumulative error effect. The error grows very large for LSTMs as we 
    predict farther into the future. The proposed method is more accurate for 
    long term predictions.
  }
  \label{fig:balltraj_example}
\end{figure*}

In this paper, we propose a novel method for trajectory prediction that mixes 
the power of deep learning and conditional generative models to provide a 
data-driven approach for accurate trajectory forecasting with the low latency 
required by real time applications. We follow a similar approach to 
conditional variational auto-encoders~\cite{cvae}, using a latent variable~$\zv$
to represent an entire trajectory, as well as an encoder and decoder network to map
trajectories to and from the latent representation~$\zv$. Our model is trained to 
maximize the conditional log-likelihood of the future observations given the past 
observations, using stochastic gradient descent and reparametrization for the 
optimization~\cite{kingma2013auto} of the variational objective. 
In addition, we introduce strategies to make the model robust to missing 
observations and outliers. We evaluate the proposed approach on a robot table tennis
setup in simulation and in the real system, showing a higher prediction accuracy
than a LSTM recurrent neural network~\cite{hochreiter1997long} and a physics-based 
model~\cite{chen2015robust}, while achieving real time execution performance. An
open-source implementation of the method presented in this paper is 
provided~\cite{dcgm_impl}.
Figure~\ref{fig:robot}, shows an image of the robot system used in the experiments,
consisting of two Barrett WAM robot arms capable of high speed movement, and
a vision system~\cite{my_vision} using four cameras with a frequency of 180 frames 
per second.

\section{TRAJECTORY PREDICTION}

The term trajectory is commonly used in the robotics community to refer
to a realization of a time series or Markov decision process. Formally, we
define a trajectory~$\tau_n = \{\obs{t}^n\}_{t=1}^{T_n}$ of total
length~$T_n$ as a sequence of multiple observations~$\obs{t}^n$, where
the index~$t$ represents time and~$n$ indexes the different trajectories
in the data set. For example, for the table tennis ball prediction problem, the 
observation~$\obs{t}^n$ is a 3 dimensional vector representing the ball position at 
time index~$t$ of the ball trajectory~$n$.

Each trajectory~$\tau_n \sim P(\tau)$ is assumed to be independently
sampled from the trajectory distribution~$P(\tau)$.
For trajectory prediction, we need to be able to predict the future 
trajectory based on previous observations. Let us use~$\yv{t}$ to denote the random 
variable representing the observation indexed by time~$t$ in any trajectory. Formally,
the goal of trajectory forecasting is to compute the conditional 
distribution~$p(\yv{t},\dots,\yv{T} \given \yv{1},\dots,\yv{t-1})$, representing the
distribution of the future values of a trajectory~$\{\yv{t},\dots,\yv{T}\}$ given 
the previous observations~$\{\yv{1},\dots,\yv{t-1}\}$. From this point on, 
we will use~$\yv{1:t}$ to denote the set of variables~$\{\yv{1},\dots,\yv{t}\}$
compactly.

Trajectory or time series forecasting methods is an active research area of machine
learning. Examples of popular approaches for time series forecasting include 
recurrent neural networks~\cite{pixelrnn}, auto-regressive 
models~\cite{cnnforecasting, wavenet} and state space models~\cite{chiappa2009using}. 
Some of which include real time performance considerations~\cite{nikhil2018convolutional}.
All these approaches share in common that 
they model~$p(\yv{t} \given \yv{1:t-1})$, and use the factorization
property of probability theory
\[ p(\yv{t:T} \given \yv{1:t-1}) = \prod_{i=t}^{T}{p(\yv{i} \given \yv{1:i-1})}, \]
to model and predict the entire future trajectory from past observations. Note
that these models predict directly only one observation into the future~$\yv{i}$ 
given the past~$\yv{1:i-1}$. To make predictions farther into the future, the 
predictions of the model are fed back into the model as additional input observations.
We will call an approach for trajectory forecasting ``recursive'' if it uses its own 
predictions as input to predict farther into the future.

\begin{comment}
The auto-regressive models learn the distribution~$p(\yv{t} \given \yv{1:t-1})$
directly. Often, only the last~$p$ observations are used to make a 
prediction~$\yv{t} = f(\yv{t-p:t-1})$. Early examples of these kind of models were 
linear~$\yv{t} = \trans{\vect{\omega}}\yv{t-p:t-1}$, but recent advances in
neural networks enabled the creation of complex non-linear auto-regressive
models with successful application to speech synthesis~\cite{wavenet} and 
dynamical systems~\cite{cnnforecasting}.

Some other models like recurrent neural networks and state space models use
a latent variable~$\vect{z}_t$ to encode all the relevant information 
of~$\yv{1:t}$, using a separate observation distribution~$p(\yv{t} \given \vect{z}_t)$ 
and an encoding distribution~$p(\vect{z}_{t+1} \given \vect{z}_{t}, \yv{t})$ to move
forward in time.
\end{comment}

An advantage of the recursive approaches is that they can model sequences
of arbitrary length by design. It is always possible to make predictions
with any given number of observations for any arbitrary number of time steps into
the future. 
On the other hand, the recursive approaches have the disadvantage that
errors are cumulative. Note that the predictions of the recursive approaches are
fed back into the model. As a result, early small prediction errors can cause
big forecasting errors as we try to predict farther into the future. For problems
with high stochasticity like traffic~\cite{yu2017long}, weather or 
stock market price prediction~\cite{cnnforecasting}, where some of these models 
are commonly applied, it is reasonable to assume that no method will ever
make almost exact long term predictions based only in previous observations. 

However, for trajectory prediction in physical systems, where we are measuring 
all the relevant variables, we would expect long term prediction to be more 
accurate. For example, we know that the model used to generate the table tennis ball 
trajectories in simulation is deterministic once the initial state is set. However,
the long term prediction error using an LSTM~\cite{hochreiter1997long} recurrent
neural network is about twice as large as using the physics-based model. 
Figure~\ref{fig:balltraj_example} shows an example ball trajectory and the
model predictions using an LSTM, depicted in green. The cumulative error 
effect for the LSTM 
model is easy to notice, specially in the Y coordinate, where the predictions 
deviate early from the ground truth ball trajectory depicted in red.

\section{DEEP CONDITIONAL GENERATIVE MODELS FOR TRAJECTORY FORECASTING}

\begin{comment}
\begin{itemize}
    \item We can also model~$p(\yv{t:T} \given \yv{1:t-1})$ directly modeling it as a regression problem.
    \item The problem: How to deal with a variable number of inputs and outputs, missing observations?
    \item We solve this problem by fixing the input and output sizes to the maximum possible value. 
      Subsequently, we pad the unavailable inputs with zeros and provide the mask to the model.
    \item We use mean squared error (MSE) as a loss function for the regression problem. For the ball prediction
      problem the MSE loss function means we want to minimize the distance between the predicted ball
      and the ground truth ball trajectory.
    \item Make a figure showing an example with many observations and other with very few observations.
      Note that the prediction accuracy is good, mention it is better than all other models discussed
      so far as shown in the experiment section. When the number of given observations is very low,
      as expected, the prediction performance is bad. We need to model uncertainty to quantity with
      the model the expected accuracy of our predictions instead of only measuring the mean behavior.
\end{itemize}
\end{comment}

We have discussed how recursive methods like recurrent neural networks suffer from cumulative
errors that render long term predictions less accurate. Therefore, our goal is to find a way
to represent the conditional distribution~$p(\obs{t:T} \given \obs{1:t-1})$ directly, in a way
where the model predictions are not fed back into the model. In addition, we want to use a powerful
model that can capture non linear relationships between the future and the past observations.

\begin{comment}
To deal with an arbitrary number of input observations~$t$, we could train~$T$ different regression models
%
\[ \{p(\obs{2:T} \given \obs{1}), p(\obs{3:T} \given \obs{1:2}), \dots, p(\obs{T} \given \obs{1:T-1})\} \]
%
for all possible number of input observations, using at test time the model for the particular
number of observations~$t$ given as input. Such an approach would not be able to exploit the 
correlation and redundancy between all the different regression models that are
performing so similar tasks. In addition, this approach requires to train and store a large number
of models.
\end{comment}

\subsection{Deterministic Regression Using Input Masks}

\rev{
Note that for a fixed value~$t$, we could model~$p(\obs{t:T} \given \obs{1:t-1})$ directly as a regression
problem. If we use a complex non-linear regression model such as a neural network, we can capture
non linear relations between the past and future observations. To deal with a variable number
of inputs~$t$ and outputs~$T-t$, 
}
we use
two auxiliary input variables~$\invar^t$ and~$\inmask^t$ that represent a zero-padded input
observations and an observation mask. Given a set of observations~$\obs{1:t-1}$ we 
set~$\invar_{1:t-1}^t = \obs{1:t-1}$,
$\invar_{t:T}^t = \vect{0}$, $\inmask_{1:t-1}^t = 1$ and~$\inmask_{t:T}^t = 0$. The 
variable~$\invar^t$, represents the observations seen so far, padding the non-observed 
part of the trajectory with zeros. 
Similarly, the variable~$\inmask^t$ represents a \{0,1\} mask indicating which
values were observed and which values were not. Using the auxiliary variables~$\invar^t$ 
and~$\inmask^t$ we can make predictions with any number of input 
observations~$t \in \{0,1,\dots,T\}$ using a single regression model even if we
have missing observations. 

The proposed approach assumes a fixed maximum prediction horizon~$T$ for all trajectories. This is a 
limitation for our approach compared to all the recursive models, that can model trajectories of any 
duration. We trade the flexibility of being able to model trajectories of arbitrary
duration for higher accuracy in the predictions and faster computation times.
\rev{In Section~\ref{sec:multiblock}, we discuss ideas to mix the power of the proposed
method with recursive approaches to be able to make predictions of any arbitrary duration.}
%We argue that for many real time applications, the benefits of high accuracy predictions and 
%fast computation times out-weight the drawback of fixing a duration horizon of interest.
%For example, in robot table tennis we need accurate ball trajectory predictions from the moment
%the opponent hits the ball to the moment it is on the reach of the robot. Therefore,

\subsection{Capturing Uncertainty and Variability}

\begin{figure}
  \centering
  \setlength\figurewidth{3.5cm}
  \setlength\figureheight{2cm}
  \begin{subfigure}[b]{0.22\textwidth}
    \resizebox{\figurewidth}{\figureheight} {
\begin{tikzpicture}
\tikzstyle{main}=[circle, minimum size = 10mm, thick, draw =black!80, node distance = 16mm]
\tikzstyle{param}=[text width=1cm, text centered, node distance = 16mm]
\tikzstyle{connect}=[-latex, thick]
\tikzstyle{box}=[rectangle, draw=black!100]
  \node[param] (mean_z) {$\vect{\mu}_z$};
  \node[param, below of=mean_z] (sigma_z) {$\vect{\sigma}_z$};
  \node[box, minimum size=10mm, left=of {$(mean_z)!0.5!(sigma_z)$}] (encoder) {$\gf$};
  \node[param, left of=encoder] (past) {$(\invar^t,\inmask^t)$};
  \path (past) edge [connect] (encoder)
        (encoder) edge [connect] (mean_z)
        (encoder) edge [connect] (sigma_z);
\end{tikzpicture}
}
    \caption{Encoder network}
    \label{fig:enc_dec:enc}
  \end{subfigure}
  \begin{subfigure}[b]{0.22\textwidth}
    \resizebox{\figurewidth}{\figureheight} {
\begin{tikzpicture}
\tikzstyle{main}=[circle, minimum size = 10mm, thick, draw =black!80, node distance = 16mm]
\tikzstyle{param}=[text width=1cm, text centered, node distance = 16mm]
\tikzstyle{connect}=[-latex, thick]
\tikzstyle{box}=[rectangle, draw=black!100]
  \node[param] (past) {$(\invar^t,\inmask^t)$};
  \node[param, below of=past] (z) {$\vect{z}$};
  \node[box, minimum size=10mm, right=of {$(past)!0.5!(z)$}] (decoder) {$\ff$};
  \node[param, right of=decoder] (future) {$\yest$};
  \path (past) edge [connect] (decoder)
        (z) edge [connect] (decoder)
        (decoder) edge [connect] (future);
\end{tikzpicture}
}
    \caption{Decoder network}
    \label{fig:enc_dec:dec}
  \end{subfigure}
  \caption{
    Encoder and decoder networks for the proposed approach. The encoder network takes 
    as input the past observations encoded in the variables~$(\invar^t,\inmask^t)$
    and produces a Gaussian distribution for the latent
    variable~$\zv$ with mean~$\vect{\mu}_z$ and standard deviation~$\vect{\sigma}_z$.
    The decoder network takes a sample~$\zv$ and the past observations and produces
    a trajectory estimate~$\yest$ including both the future~$\yest_{t:T}$ and the 
    past~$\yest_{1:t-1}$.
  }
  \label{fig:enc_dec}
\end{figure}
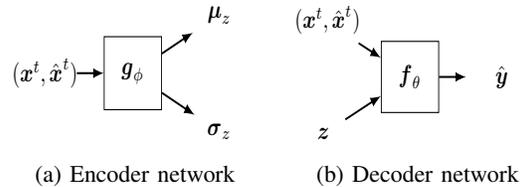

Quantifying the uncertainty of the trajectory predicted by the model is important
for decision making. A self-driving car, for example, could reduce the speed if there
is high uncertainty about the trajectory of a pedestrian crossing the street.
For real time systems, the ability to quantify uncertainty allows the agent 
to make decisions that compromise between accuracy and time to react. In
robot table tennis, for example, the robot could wait for more ball observations
if there is high uncertainty about the ball trajectory, but waiting too long will
result in failure to hit the ball.

%For a given number of observations~$\obs{1:t-1}$, there might be many possible future 
%trajectories~$\obs{t:T}$ that are likely to happen. % TODO: Show figure
%Specially, when the number of given observations is low.

\rev{
We capture uncertainty about the predictions of a trajectory~$\tau_n$ using a latent 
variable~$\zv^n$}, that can be mapped to a trajectory using a complex non-linear function, similarly to other
deep generative models approaches~\cite{vae_tutorial} like variational auto-encoders.
\rev{
We assume that the future observations~$\obs{t:T_n}^n$ are independent given the latent
variable~$\zv^n$ and the previous observation~$\obs{1:t-1}^n$, and are distributed by}
\begin{equation}
  p(\obs{t:T}^n \given \obs{1:t-1}^n, \zv) = \prod_{i=t}^{T_n} \normal{\obs{i}^n}{\yest_i^n}{\matr{\Sigma}_y},
  \label{eq:pred_model}
\end{equation}
where~$\yest^n$ is the estimated trajectory produced by the decoder network~$\ff$ 
and~$\matr{\Sigma}_y$ represents the observation noise \rev{learned also from
data}.

\rev{
  We want to emphasize that the limitation of a fixed prediction horizon~$T$ means that
  we do not have a principled approach to make predictions beyond~$T$, but we can train
  our model with trajectories of any length~$T_n$.
  If a trajectory~$\tau_n$ with length~$T_n < T$ is sampled in the training 
  mini-batch, the model still predicts a trajectory of length~$T$ but the predictions
  with~$t>T_n$ are not ``penalized'' as can be seen in~\eqref{eq:pred_model}. Note that the
  likelihood is evaluated for observations until~$T_n$. If on the other hand~$T_n \ge T$, 
  we cut a random time interval~$[t_a,t_b]$ such that~$T = t_b - t_a$ for that particular 
  mini-batch. When the same trajectory is drawn in a mini-batch later in the training process,
  a different random time interval~$[t_a,t_b]$ is used. 
  The training procedure is explained with greater detail in Section~\ref{sec:dcgm:train},
  the main message of this paragraph is that we can train our model with a data set of trajectories
  of any length. In Section~\ref{sec:multiblock}, we mention possible ideas to make predictions 
  beyond the time horizon~$T$ by mixing the advantages of recursive approaches with the method 
  presented on this paper.
  We will drop the trajectory index~$n$ from the notation from this point forward to
  avoid notational clutter.
}

%In practice, we use a single deep neural network~$\ff(\invar^t,\inmask^t,\zv)$ with output 
%size~$T \times D$ and use each of the rows indexed by~$i$ to compute~\eqref{eq:pred_model}.
\rev{Our approach, based on variational auto-encoders, will use an encoder and
decoder network to make predictions about the future value of a trajectory.
}
The decoder network, depicted in Figure~\ref{fig:enc_dec:dec}, takes as input 
the previous observations~$\obs{1:t-1}$ represented by~$(\invar^t,\inmask^t)$ as 
well as the latent variable~$\zv$ that encodes one of the possible future 
trajectories. The output of the decoder network~$\yest$ contains the predicted value
for the future observations~$\obs{t:T}$. We also use an encoder network~$\gf$ 
that produces the variational distribution~$q_\phi(\zv \given \obs{1:t})$. The
encoder network encodes a partial trajectory with observations~$\obs{1:t}$ to 
the latent space~$\zv$.

\subsection{Inference}

At prediction time, we typically want to draw several samples of the future trajectory 
conditioned on the previous observations~$p(\obs{t:T} \given \obs{1:t-1})$. To do so, 
we compute first the latent space distribution~$p(\zv \given \obs{1:t-1})$ by passing 
the given observations through the encoder network~$\gf$. 
Figure~\ref{fig:enc_dec:enc} shows a diagram of the encoder network. Given the previous 
observations, the encoder provides us with a mean~$\vect{\mu}_z$ and standard 
deviation~$\vect{\sigma}_z$ vector for the latent variable~$\zv$. 
Subsequently, we sample several values~$\zv^l \sim \normal{\zv}{\vect{\mu}_z}
{\vect{\sigma}_z}$. Each sample~$\zv^l$ and the previous ball observations are passed 
through the decoder network to obtain a sample future trajectory.

The inference process at prediction time is therefore very efficient. It requires a single
pass through the encoder and the decoding process for every sample~$\zv^l$ can be done
in parallel. In contrast with recurrent neural networks, the prediction process can be
easily parallelized, allowing fast execution even for relatively large deep learning models.

\begin{figure*}
  \centering
  \setlength\figurewidth{7cm}
  \setlength\figureheight{4.5cm}
  \setlength\axislabelsep{0mm}
  \begin{subfigure}[b]{0.45\textwidth}
    \input{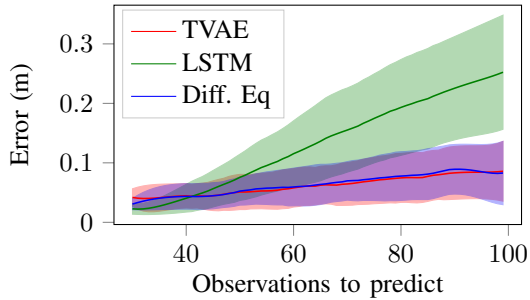}
    \caption{Prediction error in simulation}
    \label{fig:dist30:sim}
  \end{subfigure}
  ~
  \begin{subfigure}[b]{0.45\textwidth}
    \input{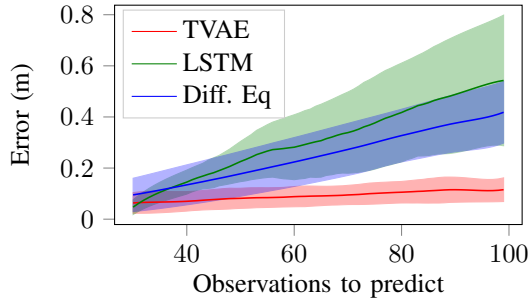}
    \caption{Prediction error in the real system}
    \label{fig:dist30:real}
  \end{subfigure}
  \caption{
    Distribution of the error in the test set for simulated data and 
    real data as a function of the number of observation to predict into
    the future. Note that in simulated data our model performs 
    as well as the differential equation based prediction, which was the 
    model used in simulation and therefore is the best we can get. 
    In real data, our model outperforms both the LSTM and differential 
    equation models, specially as we predict farther into the future.
  }
  \label{fig:dist30}
\end{figure*}

\subsection{Training Procedure}
\label{sec:dcgm:train}

The conditional likelihood using the latent variable~$\zv$ is given by
\begin{equation}
  p(\obs{t:T} \given \obs{1:t-1}) = \int{p(\obs{t:T} \given \obs{1:t-1}, \zv) p(\zv \given \obs{1:t-1}) d\zv},
  \label{eq:clh}
\end{equation}
with~$p(\obs{t:T} \given \obs{1:t-1}, \zv)$ given by~\eqref{eq:pred_model}.
We use the encoder network~$\gf$ to compute~$p(\zv \given \obs{1:t-1})$.
\rev{
In many applications, it is important to make sure the latent variable
encodes all the relevant information about the previous observations, in which
case the decoder distribution~$p(\obs{t:T} \given \obs{1:t-1}, \zv) = p(\obs{t:T} \given \zv)$.
We present the math of the model without making the previous assumption
of conditional independence for generality. Incorporating the
conditional independence assumption is trivial: simply ignore the input~$(\invar^t,\inmask^t)$
when evaluating the decoder network~$\ff$. In the experimental section, we compare the results
with and without assuming conditional independence for the table tennis ball prediction problem,
showing a slight improvement in generalization performance using the conditional independence
assumption.
}

The integral required to evaluate the conditional likelihood in~\eqref{eq:clh} is 
intractable. 
We follow the approach used for Conditional Variational Auto-Encoders~\cite{cvae} (CVAE),
optimizing instead a variational lower bound on the conditional log likelihood given by
\begin{equation}
  \begin{split}
    &\log{p_\theta(\obs{t:T} \given \obs{1:t-1})} \ge -\KL(q_\phi(\zv \given \obs{1:T}) 
    \| q_\phi(\zv \given \obs{1:t-1})) \\
    &\quad + \E_{q_\phi(\zv \given \obs{1:T})}[\log{p(\obs{t:T} \given \obs{1:t}, \zv)}],
  \end{split}
  \label{eq:elbo}
\end{equation}
where~$q_\phi(\zv \given \obs{1:T})$ is the variational distribution given by
\[
  q_\phi(\zv \given \obs{1:T}) = \prod_{k=1}^K{\normal{z_k}{\mu_z^k}{\sigma_z^k}},
\]
with~$\vect{\mu}_z$ and $\vect{\sigma}_z$ produced by the encoder network
and~$K$ is the size of the latent vector~$\zv$. 
The derivation of this objective function is presented in the 
supplementary material. 
The first term of 
the objective keeps the distributions of~$\zv$ for partial trajectory and
complete trajectories close. The second term forces the latent representation to
be a good predictor for the future trajectory. The KL divergence 
term can be computed in closed form since 
both~$q_\phi(\zv \given \obs{1:T})$ and~$q(\zv \given \obs{1:t-1})$ are Gaussian 
distributions. The expectation is approximated with Montecarlo by sampling~$\zv$ 
from the variational distribution~$q_\phi(\zv \given \obs{1:T})$.
Note that the only difference between optimizing the second term on~\eqref{eq:elbo} and 
optimizing~\eqref{eq:clh} is that the expectation is computed over a
complete or a partial trajectory respectively. This difference is key
to compute the expectation using Montecarlo. The distribution over~$\zv$ using
partial trajectories is typically too broad to be efficiently and accurately
approximated using Montecarlo, specially when the cut point~$t$ is small.

Similarly to other deep generative models like variational auto-encoders, the lower bound 
on~\eqref{eq:elbo} can be optimized using stochastic gradient descent \rev{to find the
encoder~$\vect{\phi}$ and decoder~$\vect{\theta}$ network parameters}. The 
``reparametrization trick''~\cite{kingma2013auto} is used to compute gradients. 
We provide an open source implementation of the proposed method available 
in~\cite{dcgm_impl}. The training set consists of a set of trajectories~$\tau_n$ 
each of a possibly different length~$T_n$. When we sample mini-batches to train 
our model, we randomly select a cut point~$t$ for the trajectory~$\tau_n$ 
with~$0 < t \le \tau_n$, and compute the lower bound 
for~$p(\obs{t:T} \given \obs{1:t-1})$ using the particular cut point~$t$. 
That way, our model will learn to make predictions for any number of given 
observations, including an empty set of observations. Finally, to make our 
model more robust to missing observations or outliers, we can randomly generate missing
observations and outliers for the previous observations~$\obs{1:t-1}$ in each of the 
trajectories included in the training mini-batch. To generate outliers we simply
replace an observation with a random value within the domain of the input.
\rev{
  We provide an open source implementation of the training procedure presented in this
  section on~\cite{dcgm_impl}, using Keras~\cite{keras} and TensorFlow~\cite{tensorflow}.
}

\subsection{Network Architecture}

\rev{
  The approach presented on this paper can be used with any regression method 
  for~$\ff$ and~$\gf$ as long as we can compute 
  derivatives~$\frac{d\ff}{d\vect{\theta}}$ and~$\frac{d\gf}{d\vect{\phi}}$. We used
  neural networks for this purpose in our experiments. We think that convolutional
  network architectures have great potential for time series or trajectory forecasting
  applications, since observations close in time are typically more strongly correlated
  than observations farther apart. Note that convolutional architectures have also been
  quite successful on image recognition applications, where pixels spatially close are
  typically also more strongly correlated.

  For the experiments on this paper, we only used two layer architectures with dense
  connections for simplicity. The number of neurons on the hidden layer and the
  dimensionality of the latent space~$\zv$ were selected by testing multiple powers
  of two and selecting the hyper-parameters with better performance on the validation set.
}

\subsection{Predicting Beyond the Horizon~$T$}
\label{sec:multiblock}

\rev{
  There may be many applications where making predictions arbitrarily far into the
  future is very important. Suppose for example that you use the presented approach
  to learn a forward model of a robot, and the goal is to use it to train a
  reinforcement learning agent in simulation. In such a case it is important to be
  able to make predictions much farther into the future, even at the expense of loosing
  accuracy.

  The presented approach could be extended with ideas from recursive approaches to
  make predictions far into the future. The simplest recursive idea would be to use
  our model in an auto-regressive way, as it would require no change to the math or
  software implementation. In auto-regressive mode, we would simply use the predictions
  of our model as input observations.

  A possibly better approach would be to use a state space model or recurrent neural
  network over the latent variable~$\zv$ representing block of observations. This
  approach would require to modify the encoder to receive the latent representation
  of the previous block~$i-1$ of~$T$ observations in addition to the observations seen so far
  on the current block~$i$. The encoder distribution would be therefore represented
  as~$p(\zv_i \given \zv_{i-1}, \obs{iT:iT + t})$. We do not explore any of these options
  in this paper but consider evaluating these approaches important future work.
}

\section{EXPERIMENTS}

We evaluate the proposed method to predict the trajectory of a table tennis ball in
simulation and in a real robot table tennis system. Predicting accurately the 
trajectory of a table tennis ball is difficult mostly because the spin is not 
directly observed by the vision system~\cite{my_vision}, but is significant due 
to the low mass of the ball.

We measure the prediction error and the latency, both important factors for real 
time robot applications. We use ``TVAE'' (Trajectory Variational Auto-Encoder)
to abbreviate the name of the proposed method. We compare the results with 
an LSTM and the physics-based differential equation prediction. 

For the differential equation method, we use the 
ball physics proposed in~\cite{mulling2011biomimetic}. This physics model considers 
air drag and bouncing physics but ignores spin.
To estimate the initial position and velocity, we use the approach proposed 
in~\cite{chen2015robust}, that consists on fitting a polynomial to the first~$n$ 
observations and evaluating the polynomial of degree~$k$ and its derivative
in~$t=0$. \rev{We selected~$n=30$ and~$k=2$ that provided the highest predictive
performance on our training data set.}

\rev{
To measure the prediction latency, we used a Lenovo Thinkpad X1 Carbon with an 
Intel 4 Core i7 6500U CPU of 2.50GHz and 8 GB of RAM memory. 
%We did not use a GPU for the latency experiments.
}

%We call our method for trajectory forecasting as ``dcgm'' that stands for 
%deep conditional generative model. We compare our method against Long Short term memory recurrent 
%neural networks ``lstm'' and a physical model based on differential equations.

\subsection{Prediction Accuracy in Simulation}

The simulation uses the same differential equation we used on the physics-based model. 
The results should be optimal for the physics-based model on simulation, where the 
only source of error is the initial position and velocity estimation from noisy
ball observations. To simulate the average position estimation error of the vision
system~\cite{my_vision} in simulation, we added Gaussian white noise with a standard 
deviation of 1 cm.

We generated 2000 ball trajectories for training, and another 200 for the test set. 
Figure~\ref{fig:dist30:sim} shows the prediction 
error (mean and standard deviation) in simulation over the test set. Note that 
the error distribution of the proposed method and the physics-based model is almost
identical, which is remarkable provided that the physics model used for the simulator
and the differential equation predictor are the same. The results of the LSTM are
slightly better than the proposed model for the first 10 observations into the future, but
the error for long term prediction is about three times as large as the error
for the physics or the proposed model. 

\begin{figure}
  \centering
  \setlength\figurewidth{7cm}
  \setlength\figureheight{4.5cm}
  \setlength\axislabelsep{0mm}
  \input{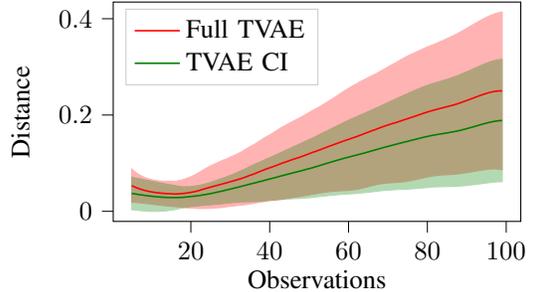}
  %\hspace{0.5cm}
  %\input{fig/avg_llh_sim}
  \caption{
    Prediction error on the test set assuming conditional independence (TVAE CI) between
    the future and the past given the latent variable~$\zv$ and without any 
    assumptions (TVAE Full). Using conditional independence forces the model to
    represent all the information about the past in the latent variable.
    Although both error curves are similar, assuming conditional 
    independence~$p(\obs{t:T} \given \zv, \obs{1:t-1}) = p(\obs{t:T} \given \zv)$
    presented a smaller average generalization error.
  }
  \label{fig:avg_err:dep_indep}
\end{figure}

\subsection{Prediction Accuracy in the Real System}

The real system consists of four RGB cameras taking 180 pictures per second attached 
to the ceiling. The images are processed with the stereo vision system proposed 
in~\cite{my_vision}, obtaining estimations of the position of the ball. There 
are several issues that make ball prediction harder on the real system: There are missing 
observations, the error is not the same in all the space due to the effects of lens 
distortion, and the ball spin can not be observed directly. We used the vision system
to collect \rev{a data set of ball trajectories including as much variability as possible,
throwing balls with the hand, with a mechanical ball launcher, and hitting them with a
table tennis racket. We randomly permuted the collected ball trajectory order and subsequently
selected the first 614 trajectories for the training set and 35 trajectories for the test set. 
The training algorithm further splits the training set into a 90\% for actual training and a remaining
10\% for the validation set used to optimize the model hyper-parameters. For both our model and the 
LSTM, we used latent variable~$\zv$ of size 64, which was the power 
of two values with better validation performance.}
\rev{In the supplementary material and in our software repository~\cite{dcgm_impl}, 
you will find the data sets collected and used for this experiments, along with a 
Python script to plot a small subset of trajectories.
The trajectories have typically a duration between 0.8 and 1.2 seconds. We used
a time horizon of $T=1.2$ seconds for our experiments.}

\rev{
First, let us compare the generalization error of the presented model with and without
making the conditional independence assumption~$p(\obs{t:T} \given \zv, \obs{1:t-1}) = p(\obs{t:T} \given \zv)$.
Figure~\ref{fig:avg_err:dep_indep} shows the prediction error on the test set collected
on the real system with (TVAE CI) and without (Full TVAE) this conditional independence
assumption, given the first 30 observations. 
Assuming~$p(\obs{t:T} \given \zv, \obs{1:t-1}) = p(\obs{t:T} \given \zv)$ might
be important in many applications to ensure that the latent variable~$\zv$ encodes all the information
necessary to predict the trajectory. In the training set the accuracy of the Full TVAE has to be
better than the TVAE CI. However, notice that on the test set we obtained a slightly smaller 
average error using the conditional independence assumption.
}

Figure~\ref{fig:dist30} compares the proposed method with the physics-based
model and the LSTM. Note that in the real system, our
model outperforms the long term prediction accuracy of the other models. The
LSTM, as expected, is very precise at the beginning but starts to accumulate 
errors and becomes quickly less accurate. The physics-based system is less accurate
than the proposed model, but is more accurate than the LSTM. The reason is because the
physics model without spin is a good approximation in cases where the spin of the ball
is very low.

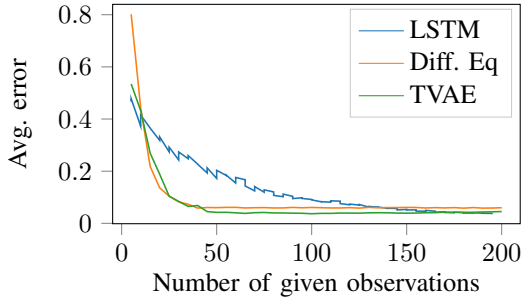
\begin{figure}
  \centering
  \setlength\figurewidth{7cm}
  \setlength\figureheight{4.5cm}
  \setlength\axislabelsep{0mm}
  % This file was created by matplotlib2tikz v0.6.14.
\begin{tikzpicture}

\definecolor{color1}{rgb}{1,0.498039215686275,0.0549019607843137}
\definecolor{color0}{rgb}{0.12156862745098,0.466666666666667,0.705882352941177}
\definecolor{color2}{rgb}{0.172549019607843,0.627450980392157,0.172549019607843}

\begin{axis}[
xlabel={Number of given observations},
ylabel={Avg. error},
xmin=-4.75, xmax=209.75,
ymin=-0.00144347285173085, ymax=0.839838074873761,
width=\figurewidth,
height=\figureheight,
tick align=outside,
tick pos=left,
x grid style={lightgray!92.02614379084967!black},
y grid style={lightgray!92.02614379084967!black},
legend style={draw=white!80.0!black},
legend entries={{LSTM},{Diff. Eq},{TVAE}},
legend cell align={left}
]
\addlegendimage{no markers, color0}
\addlegendimage{no markers, color1}
\addlegendimage{no markers, color2}
\addplot [semithick, color0]
table {%
5 0.466526768312655
5 0.477803857652829
10 0.37256468356178
10 0.415295803059587
15 0.365357319543057
15 0.364993045918129
20 0.318257273640818
20 0.331415723938522
25 0.274095474368704
25 0.292154495478007
30 0.243062265611139
30 0.273756155083439
35 0.247821186155293
35 0.259852429599434
40 0.227301737223257
40 0.22671590268678
45 0.193101316425061
45 0.212672390838207
50 0.172230656511774
50 0.203251825240038
55 0.183882493020407
55 0.185050568369021
60 0.154594617356411
60 0.176020207747772
65 0.167324236979676
65 0.143335025578599
70 0.12425970248402
70 0.140831963079126
75 0.119403702362819
75 0.12831059061421
80 0.120508158496593
80 0.107129017848981
85 0.102073826123479
85 0.112048032378857
90 0.103569645726835
90 0.0948474544003241
95 0.0980684096631739
95 0.0953112311319119
100 0.0916350168194597
100 0.0899381225394842
105 0.0826066226408653
105 0.0830862749961788
110 0.0811137522059062
110 0.0863087802165117
115 0.0851080136977284
115 0.0756674897196851
120 0.0710309565245827
120 0.0737766841376797
125 0.0699584625554293
125 0.0722899484097329
130 0.0623009823367333
130 0.0648758538963096
135 0.061971452883909
135 0.0625766504472881
140 0.0570952485526423
140 0.0579901827784515
145 0.0567794005690407
145 0.0514357501491567
150 0.05343498514706
150 0.0513249262583916
155 0.0515653723353896
155 0.0468229026956889
160 0.0463163148204149
160 0.0496289335130902
165 0.0465295472126231
165 0.0427466464784415
170 0.0443761275809994
170 0.0415163605638281
175 0.0404252468096912
175 0.0426075267695011
180 0.0408698943574131
180 0.039276868185204
185 0.0393198333685118
185 0.0401845503992319
190 0.0392641665898294
190 0.03856744679383
195 0.0380484110494991
195 0.0389873289174684
};
\addplot [semithick, color1]
table {%
5 0.801598004522602
10 0.443979062942202
15 0.218266584946182
20 0.13622162462079
25 0.103949858836081
30 0.0824722870042878
35 0.0730417091627128
40 0.0584894136535902
45 0.0606591015411612
50 0.0601718118436253
55 0.061184234299101
60 0.0611469512283535
65 0.0590478173648771
70 0.0598715584216215
75 0.0603145822892817
80 0.0589529872161601
85 0.0590835034919871
90 0.0610683309453187
95 0.059668564716415
100 0.0608613645991992
105 0.0599004651945958
110 0.0590493475741743
115 0.0589947629223732
120 0.060189167914092
125 0.058360012460874
130 0.0611824900355118
135 0.0595693733906434
140 0.0605303762110693
145 0.0603918176822631
150 0.0610457687400598
155 0.0607458061900642
160 0.0588656630086061
165 0.060165940348818
170 0.0590755739828756
175 0.0603529331175185
180 0.0581966701015251
185 0.0604859255045371
190 0.0583201462509452
195 0.0586690276539067
200 0.0599118093883096
};
\addplot [semithick, color2]
table {%
5 0.534638180927905
10 0.432379754589411
15 0.269068402692435
20 0.188529640688609
25 0.10403475316789
30 0.0839787194396295
35 0.0647936416414063
40 0.068448721401925
45 0.0450089139017154
50 0.0420483996802211
55 0.0422851915987096
60 0.040412767262273
65 0.0380671009371752
70 0.0408667798179794
75 0.0416607716195693
80 0.0398646394081739
85 0.0397118409565671
90 0.039036609805333
95 0.0385088211395875
100 0.0367965974994279
105 0.0382275306006904
110 0.0381373947636909
115 0.0394347963843194
120 0.0392191855892934
125 0.0389688991529026
130 0.0402940791883647
135 0.0407351394279478
140 0.0404423064345381
145 0.0391879775825546
150 0.0389378240392142
155 0.0393969829719619
160 0.0405033314721264
165 0.0409566207016109
170 0.0427826508029409
175 0.0427935773587065
180 0.0422563431354097
185 0.0424532895073371
190 0.0434241239692109
195 0.0441224109988856
200 0.0452044486438256
};
\end{axis}

\end{tikzpicture}
  %\hspace{0.5cm}
  %\input{fig/avg_llh_sim}
  \caption{
    Prediction error and likelihood on simulated data as a function of the number 
    of given observations. The error for the proposed model and the physics-based
    model converge with between 30 and 50 observations, whereas the LSTM model needs
    between 100 and 150 observations to obtain a similar error rate.
  }
  \label{fig:avg_err:sim}
\end{figure}

\subsection{Number of Input Ball Observations}

The trajectory prediction task consists on estimating the future trajectory~$\yv{t:T}$
given the past observations~$\yv{1:t-1}$. Accurate predictions with a relatively low 
number of input observations~$t$ is important to allow for reaction time for the robot.
Figure~\ref{fig:avg_err:sim} shows the average prediction error over the entire ball 
trajectories as we vary the number of input observations~$t$. Note that the prediction
error converges for the proposed model with~$t$ between 40 and 50 observations.
Similarly for the physics-based model. On the other hand, the LSTM error converges
after approximately 150 input observations, allowing a very low reaction time for
the robot.

\subsection{Robot Table Tennis}

The robot table tennis approach presented in~\cite{gomez2018adaptation} consists
of learning a Probabilistic Movement Primitive (ProMP) from human demonstrations,
and subsequently adapt the ProMP to intersect the trajectory of the ball. To
use~\cite{gomez2018adaptation}, the trajectory of the ball must be
represented as a probability distribution for three main reasons:
First, the initial time and duration of the movement
primitive are computed by maximizing the likelihood of hitting the ball. Second,
the movement primitive is adapted to hit the ball using a probability
distribution by conditioning the racket distribution to intersect the ball
distribution. Third, to
avoid dangerous movements, the robot does not execute the ProMP if the likelihood is
lower than a certain threshold. All these operations would not work if only
the mean ball trajectory prediction is available. 

We modified the ProMP based policy to use the proposed ball model. To compute
the ball distribution we took 30 trajectory samples from our model and computed
empirically its mean and covariance. We obtained a hitting rate 
of~\textbf{98.9\%} compared to a~\textbf{96.7\%} reported in~\cite{gomez2018adaptation} 
obtained using a ProMP as well for the ball model.

One important difference between the adapted table tennis policy 
and~\cite{gomez2018adaptation} is that we do not need to retrain the ball model 
every time the ball gun position or orientation changes. Using a ProMP as a ball 
model is only accurate if all the trajectories are very similar. Whereas our
approach can accurately predict ball trajectories with high variability.
This experiment also shows that the presented approach can be used in a
system with hard real time constraints. Our system can infer the
future ball trajectory from past observations with a latency between~\textbf{8 ms}
and~\textbf{10 ms}.

\section{Conclusions}
\label{sec:conclusions}

This paper introduces a new method to make prediction of time series 
with neural networks. We use a Gaussian distributed latent variable that encodes 
different trajectory realizations, allowing us to draw trajectory samples
from the learned trajectory distribution conditioned on any arbitrary number 
of previous observations. The proposed method is suitable for real time 
performance applications such as robot table tennis. 
We discussed why our method does not suffer from the 
cumulative error problem that popular time series forecasting methods such as 
LSTM have, and showed empirically that our method provides better long term 
predictions than other competing methods on a ball trajectory prediction
task.

\bibliographystyle{plain}
\bibliography{refs}

\end{document}